\documentclass[11pt,a4paper]{article}
\usepackage[hyperref]{emnlp-ijcnlp-2019}
\usepackage{times}
\usepackage{latexsym}
\usepackage{amsmath}
\usepackage{url}
\usepackage{multirow}

\title{Transformer-Based Conditioned Variational Autoencoder for Dialogue Generation}

\author{Huihui Yang \\
  Zhejiang University \\
  {\tt yanghh0@zju.edu.cn}
  }

\date{}
\usepackage{graphicx}

\begin{document}
\maketitle
\begin{abstract}
In human dialogue, ang one query usually elicits numerous appropriate responses.
The Transformer-based dialogue model produces frequently occurring sentences in the corpus since it is a one-to-one mapping function.
CVAE is a technique for reducing generic replies.
In this paper, we create a dialogue model (CVAE-T) based on the Transformer with CVAE structure.
We use a pre-trained MLM model to rewrite some key n-grams in responses to obtain a series of negative examples, and introduce a regularization term during training to explicitly guide the latent variable in learning the semantic differences between each pair of positive and negative examples. 
Experiments suggest that the method we design is capable of producing more informative replies.
\end{abstract}

\section{Introduction}
The training data used to train the dialogue models contains a great deal of unknown background information, making the dialogue a one-to-many problem where different people can come up with different but reasonable answers to the same question.
Generative diversity is a crucial characteristic for building dialogue systems.
\citealt{zhao2017learning} use CVAE for dialogue modeling and demonstrate that the sentences produced by CVAE model are more diverse than those produced by conventional sequence-to-sequence model.

For CVAE model, the approximate posterior carries little useful information at the beginning of training, and the model tends to fit the distribution directly without reference to the latent variable, which is known as the KL-vanishing~\cite{bowman2015generating}. 
In order to alleviate this problem, some researchers introduce dialogue intent labels~\cite{zhao2017learning} or sentence function labels (interrogative, declarative and imperative)~\cite{ke2018generating} as additional information to supervise the posterior network learning.
However, this method has many drawbacks: 1) It is expensive to annotate labels and challenging to expand to large-scale datasets. 2) It only focuses on the attributes of a certain aspect of sentences, and the limited number of tags are difficult to cover all the attributes of that aspect. 3) The tags themselves do not carry semantic information, which is not conducive to model learning.
We observe that some key words or phrases in the sentence can serve as representations of high-level sentence attributes, disregarding the need for additional tags.
We locate the key n-grams in each response using a keyword extraction algorithm and replace them with a special token [MASK] respectively.
These masked positions are rewritten by a pre-trained MLM model to generate a series of negative sentences semantically distinct from the original sentence.
A regularization term is used to constrain the prior and posterior distribution during training, helping the latent variable to perceive the difference between positive and negative examples.

Dialogue models should be able to handle long dependencies well because conversation datasets usually contain multiple rounds of sentences, and as the conversation goes on, the dialogue history accumulates into a very long sequence.
Transformer-based models~\cite{zhang2019dialogpt,roller2020recipes} have shown strong generative power when trained on large-scale conversational corpora.
Due to its self-attention mechanism and excellent parallelism, Transformer is suited for processing prolonged sequences.
Its hierarchical structure also enables the decoder to incorporate latent variable in a more flexible manner.
We choose Transformer as encoder-decoder framework and explore how the CVAE structure could be better integrated with it for dialogue generation.

The contributions of this paper can be summarized as follows:

\begin{itemize} 
\item
We design a Transformer-based conditioned variational autoencoder for dialogue generation, named CVAE-T.
\item
We utilize a simple and effective method to prompt the latent variable learning more meaningful distribution.
\item
Experiments illustrate that CVAE-T achieves significant improvements on diversity and informativeness for dialogue generation.
\end{itemize}

\section{Model}
\begin{figure*}[ht]
\centering
\includegraphics[width = .55\textwidth]{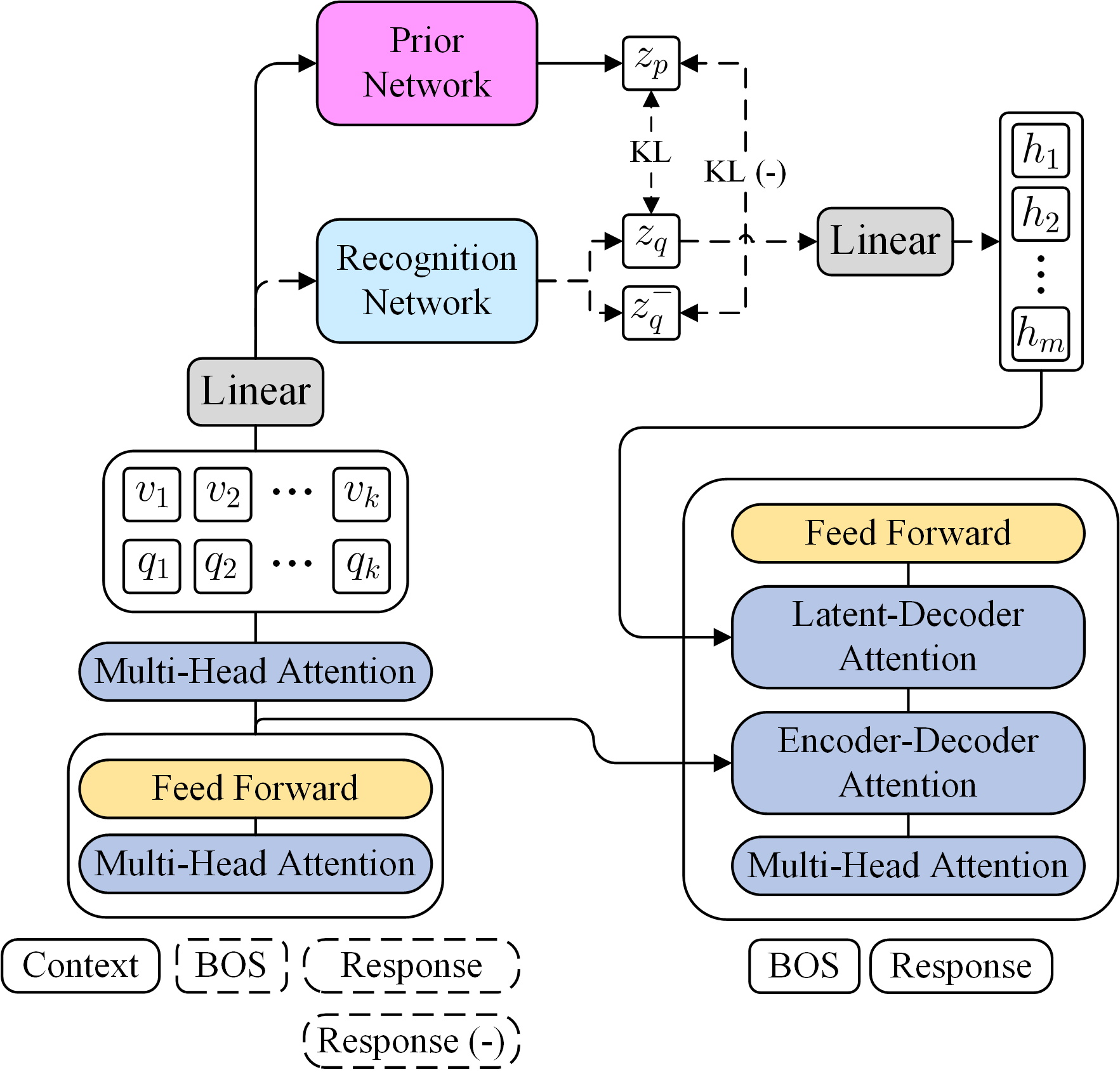}
\caption{
The framework of our model.
The latent variable fed into decoder is sampled from the recognition network connected by a dashed line during training while from the prior network during validating and inference.
}
\label{fig:framework}
\end{figure*}

In this section, we first define our task and then describe in detail the components of our designed model.
The overall architecture of our model is illustrated in Figure~\ref{fig:framework}.

\subsection{Task Definition}
In our definition of dyadic conversation task, there are three elements: dialogue context $c$, response $y$ and latent variable $z$.
Dialogue context $c$ is the concatation of conversation history $h$ and current query $x$, which can be denoted as $c=[h;x]$.
Latent variable $z$ is utilized to model the probability distribution of different potential factors influencing conversation generation.
The dialogue generation task can be expressed as the following conditional probability:
\begin{equation}
\begin{split}
p(y,z|c)=p(z|c) \cdot p(y|c,z)
\end{split}
\end{equation}
$p(z|c)$ represents the sampling process of the latent variable $z$, which is approximated by a neural network called prior network with parameter $\theta$.
$p(y|c,z)$ represents the decoding process and is approximated by the decoder network in the encoder-decoder framework.

\subsection{Input Representation}
\begin{figure}[ht]
\centering
\includegraphics[width = .35\textwidth]{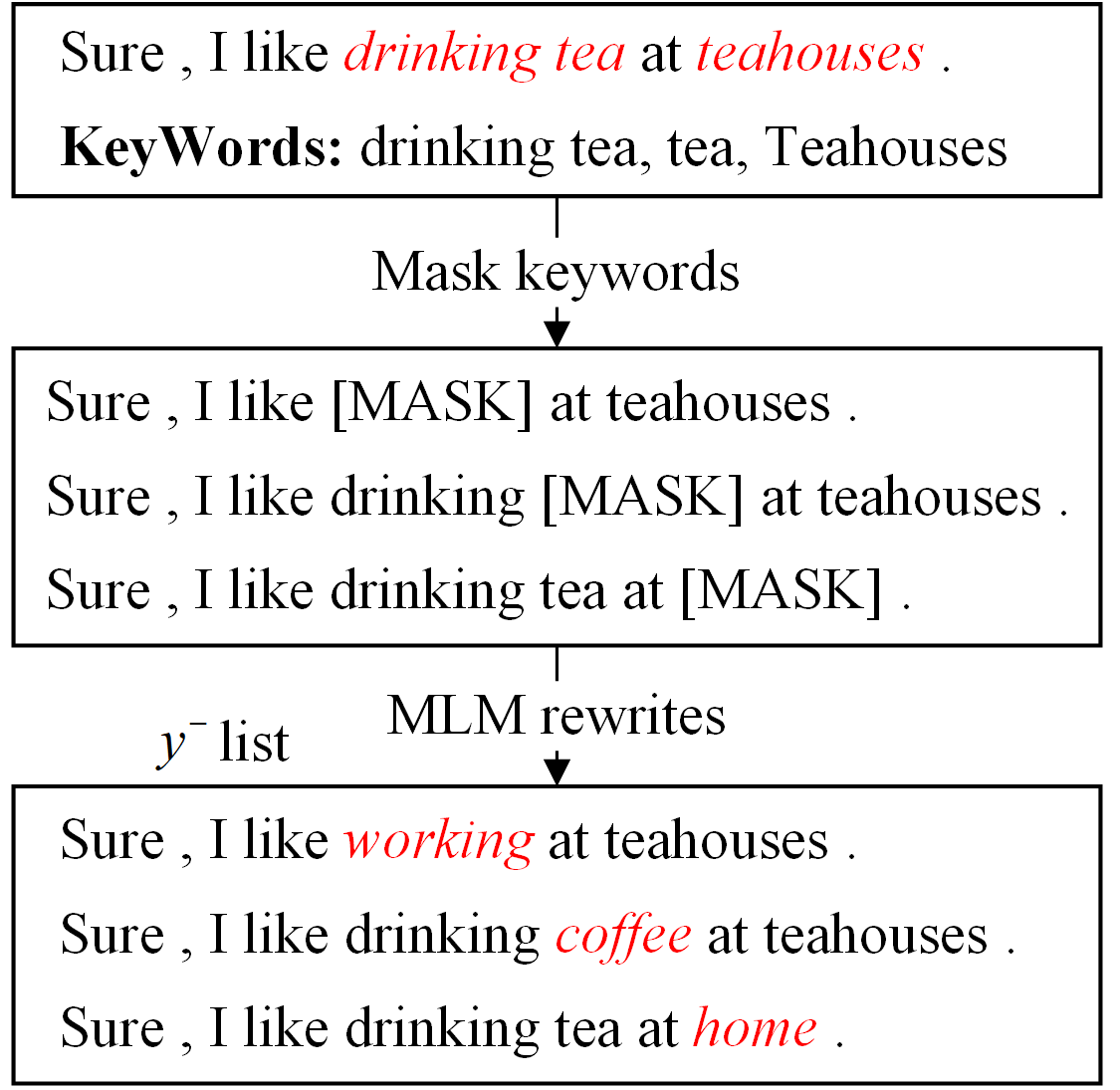}
\caption{Negative sentences generation process.}
\label{fig:regular}
\end{figure}

\begin{figure*}[ht]
\centering
\includegraphics[width = .6\textwidth]{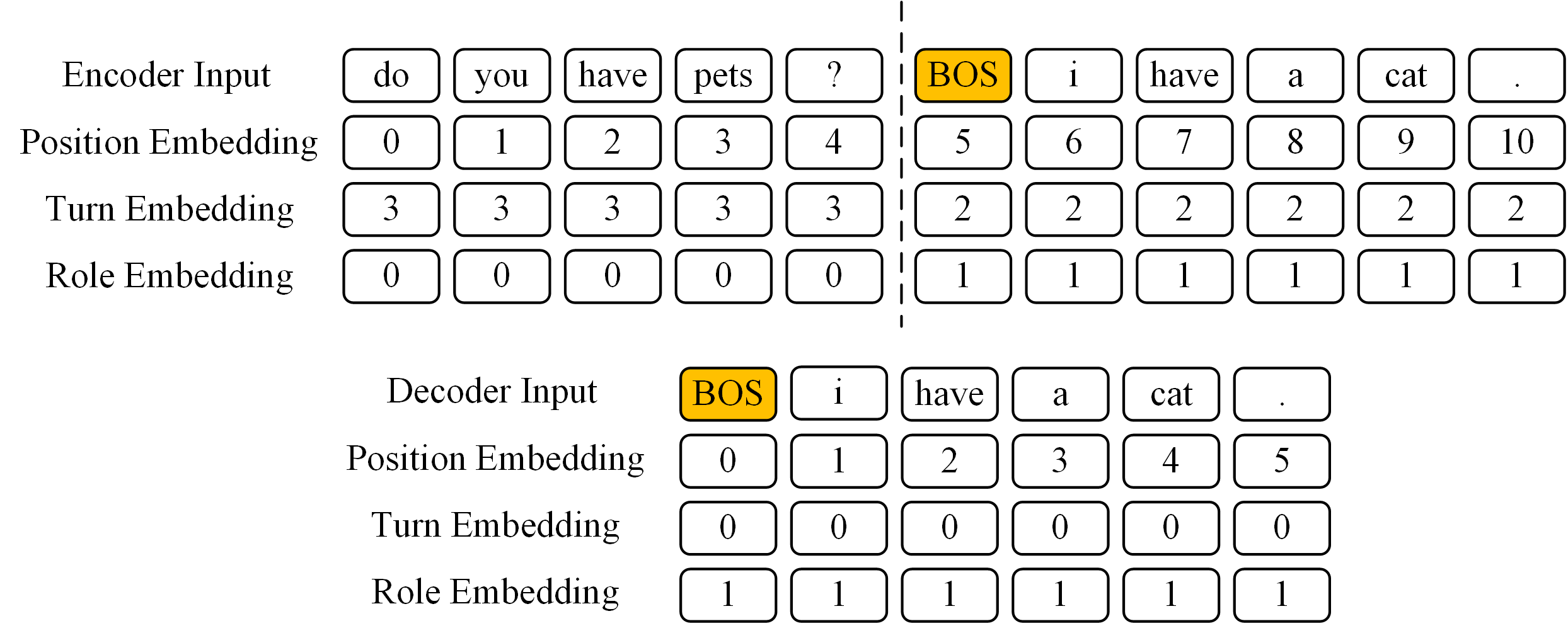}
\caption{Input representation. The token embedding is not drawn in the figure.
}
\label{fig:inputs}
\end{figure*}

Besides the word embedding and position embedding used in the original Transformer, we also employ another two kinds of embedding, turn embedding and role embedding, similar to PLATO~\cite{bao2019plato} to represent a token.
Each utterance $x$ in a dialogue context $c$ is assigned a turn id, decreasing sequentially from the maximum turns to 1, and the turn id of $y$ is always set to 0.
As there are two speakers in each dialog episode for our task, we set two role ids: 0 for the person who speaks first and 1 for the other.
The final input embedding of a token is the sum of the corresponding word, position, turn and role embedding, as shown in Figure~\ref{fig:inputs}.

\subsection{Negative Sentences Generation}
We use the Yake (Yet Another Keyword Extractor) algorithm~\cite{campos2018yake}, which is an unsupervised method, to extract the keywords for each response.
Yake creates a set of five features to capture the unique characteristics of each term. These are as follows: (1) Casing, (2) Word Positional, (3) Word Frequency, (4) Word Relatedness to Context, and (5) Word DifSentence.
Inspired by the generation of pseudo-QE data in machine translation quality evaluation task, we select BERT~\cite{devlin2018bert} to rewrite the keywords found.
If $m$ key n-grams are found in a response, we substitute those n-grams with token [MASK] respectively to create $m$ sentences. 
We then feed these $m$ sentences into BERT and sample the model output at each masked position to infill all masked sentences. 
The negative sentences generation process is shown in Figure~\ref{fig:regular}.

\subsection{Encoder}
Encoder block in our model obeys the original Transformer~\cite{vaswani2017attention}, which includes two key subcomponents: multi-head attention and feed-forward network.
Each subcomponent is accompanied with a residual connection and layer normalization.
During training, the encoder needs to be called three times to encode three different types of input data.
They are as follows: (1) dialogue context $c$, (2) the concatation of dialogue context $c$ and response $y$, (3) the concatation of dialogue context $c$ and negative response $y^-$.
$y^-$ is chosen randomly from the list of negative sentences of $y$.

\subsection{Recognition/Prior Network}
Our model utilizes a latent variable $z$ to determine the high-level semantics between $c$ and $y$. As the true posterior distribution $p(z|c,y)$ is intractable, a recognition network $q_\phi(z|c,y)$ is used to approximate it so that we can sample $z$ from $q$ during training.
Likewise, a prior network $p_\theta(z|c)$ is introduced to approximate the real prior distribution $p(z|c)$.
We adopt the assumption that $z$ follows multivariate Gaussian distribution, so $p_\theta$ and $q_\phi$ are normally distributed:
\begin{equation}
\begin{split}
p_\theta(z|c) &\sim \mathcal N(\mu_p,\sigma_p^{2}\rm{\mathbf{I}}) \\
q_\phi(z|c,y) &\sim \mathcal N(\mu_q,\sigma_q^{2}\rm{\mathbf{I}})
\end{split}
\end{equation}

We define $k$ vectors $\left \{ q_1,q_2,\cdots,q_k \right \}$ as queries.
For each query, we use attention mechanism to aggregate the output vectors of the encoder.
After repeating $k$ times, we obtain $k$ vectors and the final sentence representation is a vector transformed from a linear network, whose input is the concatenation of these k vectors.
The means and variances of $p_\theta$ and $q_\phi$ are computed as follows:
\begin{equation}
\begin{split}
\begin{bmatrix}
\mu_p\\ 
{\rm log}(\sigma_p^{2})
\end{bmatrix} &= {\rm MLP}_p(e(c)) \\
\begin{bmatrix}
\mu_q\\ 
{\rm log}(\sigma_q^{2})
\end{bmatrix} &= W_q (e(c \oplus y))+b_q
\end{split}
\end{equation}
where $e(c)$ is the representative vector of $c$ and ${\rm MLP}_{p}$ is a 2-layer multi-layer perceptron with a tanh activation.



\subsection{Decoder}
After obtaining a vector representation of the latent variable $z$, we map it into a vector group H using a linear network:
\begin{equation}
\begin{split}
H=W_h \cdot z
\end{split}
\end{equation}
To incorporate the latent variable into decoder, we introduce a Latent-Decoder Attention layer that shares weights with the Encoder-Decoder Attention layer to perform attention operation on $H$.
We denote the input and output of the $l$-th decoder block as $D_{i}^{l},D_{o}^{l}$ and denote the final output of the encoder as $E$.
Then its data flow process can be formalized as:
\begin{equation}
\begin{split}
A &= MultiHead(D_{i}^{l}, D_{i}^{l}, D_{i}^{l} ) \\
B &= MultiHead(A, E, E) \\
C &= MultiHead(B, H, H) \\
\end{split}
\end{equation}

\subsection{Objectives}
The widely adopted NLL loss is embraced in the training:
\begin{equation}
\begin{split}
L_{NLL} = -E_{z \sim q_\phi(z|c,y)}\sum_{t=1}^{|y|}p(y_t|c,z,y_{<t})
\end{split}
\end{equation}
To reduce the KL distance between the prior distribution and the positive posterior distribution and increase the KL distance between the prior distribution and the negative posterior distribution, we introduce a regularization term:
\begin{equation}
\begin{split}
KL^+ &= KL\left (q_\phi(z|c,y) \parallel p_\theta(z|c)  \right ) \\
KL^- &= KL\left (q_\phi(z|c,y^-) \parallel p_\theta(z|c)  \right ) \\
L_d &= max\left (\varepsilon,  KL^+ - KL^-\right )
\end{split}
\end{equation}
where $\varepsilon < 0$.
The bag-of-words loss~\cite{zhao2017learning} is also used to speed up the training of latent variable:
\begin{equation}
\begin{split}
L_{Bow} = -E_{z \sim q_\phi(z|c,y)}\sum_{t=1}^{|y|}p(y_t|c,z) \\
p(y_t ) \sim softmax(MLP_b (e(c) \oplus z))
\end{split}
\end{equation}
where ${\rm MLP}_{b}$ is a 2-layer multi-layer perceptron with a tanh activation.
To summarize, the overall goal of our model is to minimize the integrated loss:
\begin{equation}
\begin{split}
L = L_{NLL} + L_{Bow} + L_d
\end{split}
\end{equation}

\section{Experiments}

\subsection{Datasets}

\begin{table}[ht]
\begin{center}
\scalebox{0.8}{
\begin{tabular}{ccccc}
\hline \bf Corpus& \bf \#Train& \bf \#Valid& \bf \#Test \\ \hline
Dailydialog & 22,236 & 2,000 & 2,000 \\
Convai2 & 17,878 & 1,000 & 1,000 \\
\hline
\end{tabular}}
\end{center}
\caption{\label{font-t1} Number of dialogue episodes for each dataset.}
\end{table}

We select two publicly available datasets, DailyDialog~\cite{li2017dailydialog} and ConvAI2~\cite{zhang2018personalizing}, as benchmark datasets.
DailyDialog (DD) is a chitchat dataset which covers ten categories ranging from holidays to financial topics and can reflect conversations occurring in daily life.
ConvAI2 (CA) is a dataset biased towards persona.
Each speaker is given a role to play based on sentences describing their persona and both participants engage in dialogue around their roles in an attempt to know each other.
The overall statistics of these two datasets are summarized in table~\ref{font-t1}.

\subsection{Compared Methods}
The following models have been compared with our model.

{\bf AttnS2S}
It is a generic Seq2Seq model based on LSTM with a global attention mechanism~\cite{bahdanau2014neural}.

{\bf CVAE}
It is a CVAE model based on LSTM.
Unlike ~\cite{zhao2017learning}, the CVAE model we implement does not include any additional linguistic features to supervise latent variable learning.

{\bf Transformer}
Because our model is based on the Vanilla Transformer, it is the most appropriate baseline.

\subsection{Implementation Details}
Our implementation is built on the ParlAI framework~\cite{miller2017parlai}.
The hidden size of the RNN model is set to 512, and the decoder is configured as a 2-layer unidirectional LSTM.
The encoder of AttnS2S is a 2-layer bidirectional LSTM.
The CVAE model has a 1-layer bidirectional LSTM and a 1-layer unidirectional LSTM for encoding single sentence and dialogue context respectively.
For Transformer, we set hyper-parameters as 6-layer encoder, 6-layer decoder, 8 attention head, 512 inner hidden size and 3072 feed-forward filter size.
For VAE, the dimension of the latent variable $z$ is set to 128.
For our model, the number of query vectors $k$ is set to 4 and the number of vectors contained in $H$ transformered from $z$ is set to 6.
We also use KL annealing~\cite{bowman2015generating} (20,000 steps for each dataset) to achieve better performance.
For all models, we use GPT2 vocabulary and the embedding size is set to 512.
Another four special tokens ($\_\_unk\_\_$, $\_\_start\_\_$, $\_\_end\_\_$ and $\_\_null\_\_$) are added to the vocabulary.

The dialogue context is a sequence of all historical dialogue sentences concatenated with delimiter $\setminus n$.
The maximum sequence length of context and response is set to 360 and 72 respectively.
Adamax optimizer is employed for optimization with an initial learning rate of $1 \times 10^{-4}$.
During testing, we use standard beam search with a beam size of 10, no minimum beam decoding constraint, but with context and response 3-gram blocking~\cite{roller2020recipes}.

\subsection{Metric}

To evaluate the responses generated by all compared methods, we conduct automatic evaluation and manual evaluation on two test sets.

For automatic evaluation, we report: 
(1) {\bf Perplexity (PPL)} reflects the difficulty of generating ground truth responses. 
(2) {\bf BLEU}~\cite{chen2014systematic} measures word overlap between the generated response and the ground truth.
We use BLEU-1 to 4 as our lexical similarity metric.
(3) {\bf DISTINCT-1/2}~\cite{li2015diversity} can measure the diversity of generated responses by calculating the ratio of distinct unigrams/bigrams in all generated tokens.

For manual evaluation, we randomly collect 100 dialogue contexts with at least three conversational turns and the corresponding responses generated by all models in the two test sets.
Annotators are asked to compare the responses (win, tie or lose) quality from two aspects: coherence and informativeness.
We use Fleiss'Kappa~\cite{fleiss1971measuring} to measure agreement among the annotators.

\section{Results and Analysis}

\subsection{Main Results}
\label{4.1}
In table~\ref{font-t2}, we compare two CVAE-T variants with baseline models.
In table~\ref{font-tx}, we compare our model with the one without $l_d$ loss on the validate dataset.
The CVAE-T model outperforms the other models in DISTINCT-1/2 metrics, indicating that our model can generate more diverse responses.
This is because of the sampling operation performed at each generation, with different potential factors yielding different responses.
The BLEU and PPL scores of Transformed-based approaches is better than that of RNN-based models.
\begin{table}[ht]
\begin{center}
\scalebox{0.63}{
\begin{tabular}{c|c|c|c|c}
    \hline
    Corpus & Methods & PPL & BLEU-1/2/3/4 & DISTINCT-1/2 \\
    \hline
    \multirow{5}{*}{DD} 
    & AttnS2S & 59.1 & 8.75 / 4.23 / 2.30 / 1.38 & 2.35 / 11.95 \\
    \cline{2-5}
    & CVAE & 83.0 & 9.92 / 4.02 / 2.04 / 0.93 & 2.20 / 14.42 \\
    \cline{2-5}
    & Transformer & \textbf{30.6} & 10.73 / 5.34 / 3.16 / 2.05 & 2.60 / 15.01 \\
    \cline{2-5}
    & CVAE-T & 34.6 & \textbf{11.64} / \textbf{5.54} / \textbf{3.24} / \textbf{2.21} & \textbf{3.05} / \textbf{19.96} \\
    & $-l_d$ & 34.6 & 11.20 / 5.28 / 3.02 / 1.84 & 2.72 / 18.42 \\
    \hline
    
    \multirow{5}{*}{CA} 
    & AttnS2S & 54.7 & 10.7 / 3.86 / 1.44 / 0.60 & 1.07 / 4.63 \\
    \cline{2-5}
    & CVAE & 135.7 & 8.88 / 1.96 / 0.50 / 0.15 & 0.92 / 5.68 \\
    \cline{2-5}
    & Transformer & \textbf{36.4} & \textbf{11.87} / \textbf{4.29} / \textbf{1.67} / \textbf{0.69} & 0.88 / 3.83 \\
    \cline{2-5}
    & CVAE-T & 48.2 & 11.32 / 3.30 / 1.07 / 0.46 & 1.31 / \textbf{7.98} \\
    & $-l_d$ & 48.9 & 11.28 / 3.18 / 1.06 / 0.33 & \textbf{1.35} / 7.94 \\
    \hline
\end{tabular}}
\end{center}
\caption{\label{font-t2} Experimental results on automatic measures.}
\end{table}
\begin{table}[h!]
\begin{center}
\scalebox{0.62}{
\begin{tabular}{c|c|c}
    \hline
    Corpus & Methods & KL cost \\
    \hline
    \multirow{2}{*}{DD} 
    & CVAE-T & \textbf{5.0} \\
    & $-l_d$ & 3.6 \\
    \hline
    
    \multirow{2}{*}{CA} 
    & CVAE-T & \textbf{9.6} \\
    & $-l_d$ & 8.8 \\
    \hline
\end{tabular}}
\end{center}
\caption{\label{font-tx} KL terms on DD and CA validate sets.}
\end{table}
\begin{table}[h!]
\begin{center}
\scalebox{0.62}{
\begin{tabular}{c|cc|cc}
    \hline
    \multirow{2}{*}{Corpus} & \multicolumn{2}{c|}{Coherence} & \multicolumn{2}{c}{Informativeness} \\
    & Win & Lose & Win & Lose \\
    \hline
    \multirow{1}{*}{DD} 
    & 13\% & 9\% & 20\% & 2\% \\
    \hline
    \multirow{1}{*}{CA}
    & 15\% & 6\% & 18\% & 3\% \\
    \hline
\end{tabular}}
\end{center}
\caption{\label{font-txx} Human evaluation of CVAE-T vs Transformer on the DD and CA test sets.}
\end{table}
Both datasets contain multi-turn communication and the dialogue history grows very long as the conversation progresses.
The results of BLEU and PPL metrics show that Transformed-based approaches are better at processing long sequences and can generate sentences of higher quality.
When the $l_d$ loss is removed from CVAE-T, the BLEU, DISTINCT and KL Cost metrics all drop significantly, and the PPL metric is almost flat.
It demonstrates that the regularization term helps model to perceive the difference between positive and negative examples, making the learned distribution more reasonable and forcing the decoder to generate more coherent and precise words.
The models trained with ConvAI2 dataset get considerably lower DISTINCT scores than the models trained with Dailydialog dataset.
This is due to the fact that ConvAI2 is a highly biased dataset.
Most dialogues revolve around persona resulting in a lot of similar contexts and responses, as well as low lexical diversity.
The PPL of RNN-based CVAE model is noticeably larger than that of other models.
\begin{figure}[ht]
\includegraphics[width = .46\textwidth]{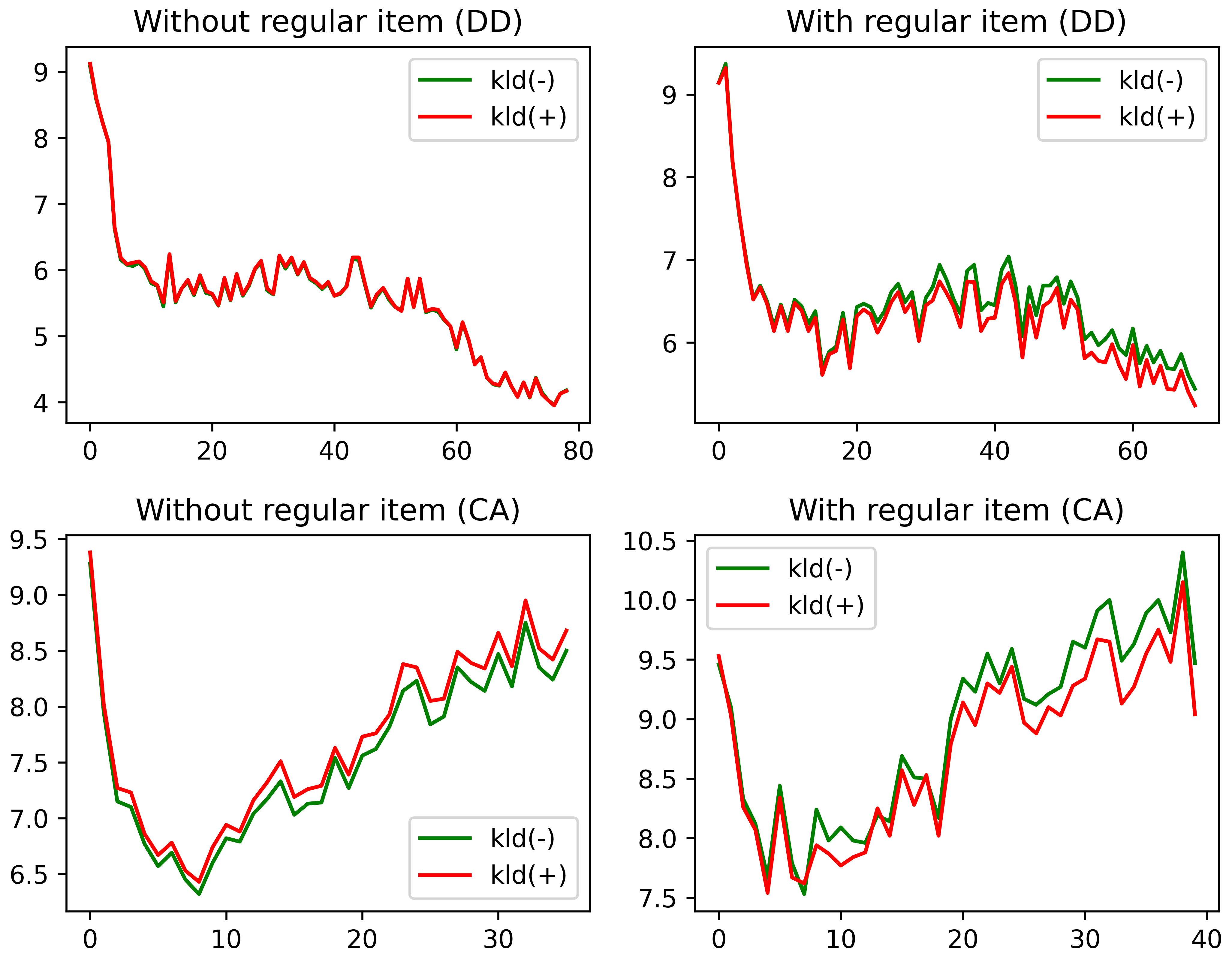}
\caption{The value of the KL divergence on two validate sets during training.}
\label{fig:constitute_tree}
\end{figure}
We analyze this phenomenon for two reasons: 
1) The encoder of CVAE is a hierarchical structure, where tokens in each sentence cannot interact directly with tokens in other sentences in the dialogue context. 
2) The way the decoder fuses latent variable $z$ is too simple.
It only uses $z$ as the initial state of the decoder RNN, so the latent variable has less control over the decoding process.
Overall, our CVAE-T obtains competitive results.

\subsection{Human Evaluation}
\label{4.2}
We conduct a human evaluation to make it more convincing. We specifically choose 100 dialogue contexts at random and generate responses using the following methods: CVAE-T and Transformer.
The results of human comparison are shown in table~\ref{font-txx}, where the average Fleiss'Kappa is 0.587 and 0.526 respectively, indicating annotators have reached moderate agreement.
In general, the two models produce comparable responses most of the time.
The CVAE-T slightly outperforms the Transformer in terms of coherence and consistency.

\subsection{Effect Of $\boldsymbol{l_d}$}
Figure~\ref{fig:constitute_tree} shows the variation of the KL divergence on two validate sets during training.
The red curve denotes the KL divergence ($KL^+$) between the prior distribution $p_\theta(z|c)$ and the positive posterior distribution $q_\phi(z|c,y)$, while the green one denotes the KL divergence ($KL^-$) between the $p_\theta(z|c)$ and the negative posterior distribution $q_\phi(z|c,y^-)$.
The reasonable situation is $KL^+$ is less than $KL^-$.
This is reflected in the figure as the red curve is below the green curve and the distance between them gradually increases as the training progresses.
However, the red curve acts similarly to or even higher than the green one when no regularization term is used.
This phenomenon shows the latent variable $z$ learns in a natural way is difficult to perceive subtle differences between sentences.
After applying the regularization term, the curve changes as expected.
This suggests that the regularization term can drive the latent variable to learn more discriminatory information.

\subsection{Case Analysis}
\begin{table*}[ht]
\begin{center}
\scalebox{0.8}{
\begin{tabular}{c|l}
    \hline
    \multirow{3}{*}{\textbf{Context}} & $\hspace{0.25cm}$ Q: How many languages can you speak ? \\
    & $\hspace{0.25cm}$ R: I can speak French and German .  \\
    & $\hspace{0.25cm}$ Q: How well can you speak them ? \\
    \hline
    GOLD & $\hspace{0.25cm}$ I can speak German quite well , but I can't speak French very well . \\
    \hline
    AttnS2S & $\hspace{0.25cm}$ I can't speak French very well . \\
    \hline
    CVAE & $\hspace{0.25cm}$ Very well , I can also speak English . \\
    \hline
    Transformer & $\hspace{0.25cm}$ I can't speak French very well . \\
    \hline
    \multirow{3}{*}{CVAE-T} & $\hspace{0.25cm}$ I can't speak French well, but I can speak German very well . \\
    & $\hspace{0.25cm}$ I can communicate with my French fluently . \\
    & $\hspace{0.25cm}$ Very well , I can also speak English . \\
    \hline
    \hline
    \multirow{3}{*}{\textbf{Context}} & $\hspace{0.25cm}$ Q: Hey Tina . What are you doing ? \\
    & $\hspace{0.25cm}$ R: I was just watching TV . What's going on with you ? \\
    & $\hspace{0.25cm}$ Q: I just watched a movie and I'm scared . \\
    \hline
    GOLD & $\hspace{0.25cm}$ What did you watch ? \\
    \hline
    AttnS2S & $\hspace{0.25cm}$ That sounds great . \\
    \hline
    CVAE & $\hspace{0.25cm}$ Really ? It ' s been a long time ! \\
    \hline
    Transformer & $\hspace{0.25cm}$ Oh , what a pity ! \\
    \hline
    \multirow{3}{*}{CVAE-T} & $\hspace{0.25cm}$ Oh , what kind of movie did you see ? \\
    & $\hspace{0.25cm}$ Me too . Do you like watching movies ? \\
    & $\hspace{0.25cm}$ Oh , what happened ? \\
    \hline
\end{tabular}}
\end{center}
\caption{\label{font-t2x} Generated responses from all baselines and our model.}
\end{table*}
We provide two generated dialogues in table~\ref{font-t2x} from the DailyDialog test set.
Each example contains three turns of historical dialogue sentences, where the speaker Q starts with a query and the speaker R generates a response.
For CVAE-T, we show three responses for each dialogue context.
Our model can produce more varied and specific responses, indicating that the latent variable does capture meaningful information.

\section{Related Work}
One approach to improving the diversity of dialogue generation model is to revise decoding strategies.
\citealt{vijayakumar2016diverse} propose a Diverse Beam Search algorithm, which decodes diverse lists by grouping the beam budget and enforcing diversity between groups of beams.
\citealt{li2015diversity} replace traditional maximum likelihood with maximum mutual information during decoding stage, allowing the decoding process to account for the correlation between inputs and outputs.
They also propose adding a penalty term to the Beam Search algorithm to penalize sibling nodes that share the same parent node, allowing the algorithm to search more different paths by expanding from different parent nodes.

Another idea to improve generative diversity is to increase the complexity of dialogue model by introducing continuous or discrete latent variable.
Our work is related to continuous latent variable.
\citealt{serban2017hierarchical} propose VHRED, a hierarchical CVAE-based sequence-to-sequence model that explicitly models the multi-level variations in the responses.
\citealt{zhao2017learning} and \citealt{ke2018generating} introduce conversational intention labels and sentence function labels (interrogative, declarative and imperative) respectively as additional linguistic features to supervise the latent variable learning.
\citealt{zhou2017mojitalk} incorporate emojis as a prior information to generate responses with specific emotions.
\citealt{wu2019guiding} use character information to control the model generating personality-perceived responses.
\citealt{gu2018dialogwae} translate the KL distance in CVAE into the Wasserstein distance of GAN and introduce a Gaussian mixture prior to support multimodality.
Recently, Transformers are considered in VAEs for storytelling~\cite{wang2019t}, classification~\cite{gururangan2019variational} and dialogue generation~\cite{li2021emoelicitor}.

\section{Conclusion and Future Work}
In this paper, we design a Transformer-based conditioned variational autoencoder for dialogue generation and utilize a simple method to prompt the latent variable learning more meaningful distribution.
One of our future work is to explore effective pre-training tasks to leverage our method.



\bibliography{emnlp-ijcnlp-2019}

\begin{thebibliography}{24}
\expandafter\ifx\csname natexlab\endcsname\relax\def\natexlab#1{#1}\fi

\bibitem[{Bahdanau et~al.(2015)Bahdanau, Cho, and Bengio}]{bahdanau2014neural}
Dzmitry Bahdanau, Kyunghyun Cho, and Yoshua Bengio. 2015.
\newblock Neural machine translation by jointly learning to align and
  translate.
\newblock In \emph{3rd International Conference on Learning Representations}.

\bibitem[{Bao et~al.(2020)Bao, He, Wang, Wu, and Wang}]{bao2019plato}
Siqi Bao, Huang He, Fan Wang, Hua Wu, and Haifeng Wang. 2020.
\newblock Plato: Pre-trained dialogue generation model with discrete latent
  variable.
\newblock In \emph{Proceedings of the 58th Annual Meeting of the Association
  for Computational Linguistics}, pages 85--96.

\bibitem[{Bowman et~al.(2016)Bowman, Vilnis, Vinyals, Dai, Jozefowicz, and
  Bengio}]{bowman2015generating}
Samuel~R Bowman, Luke Vilnis, Oriol Vinyals, Andrew~M Dai, Rafal Jozefowicz,
  and Samy Bengio. 2016.
\newblock Generating sentences from a continuous space.
\newblock In \emph{Proceedings of The 20th {SIGNLL} Conference on Computational
  Natural Language Learning}, pages 10--21.

\bibitem[{Campos et~al.(2018)Campos, Mangaravite, Pasquali, Jorge, Nunes, and
  Jatowt}]{campos2018yake}
Ricardo Campos, V{\'\i}tor Mangaravite, Arian Pasquali, Al{\'\i}pio~M{\'a}rio
  Jorge, C{\'e}lia Nunes, and Adam Jatowt. 2018.
\newblock Yake! collection-independent automatic keyword extractor.
\newblock In \emph{European Conference on Information Retrieval}, pages
  806--810. Springer.

\bibitem[{Chen and Cherry(2014)}]{chen2014systematic}
Boxing Chen and Colin Cherry. 2014.
\newblock A systematic comparison of smoothing techniques for sentence-level
  bleu.
\newblock In \emph{Proceedings of the ninth workshop on statistical machine
  translation}, pages 362--367.

\bibitem[{Devlin et~al.(2019)Devlin, Chang, Lee, and
  Toutanova}]{devlin2018bert}
Jacob Devlin, Ming-Wei Chang, Kenton Lee, and Kristina Toutanova. 2019.
\newblock Bert: Pre-training of deep bidirectional transformers for language
  understanding.
\newblock In \emph{Proceedings of the 2019 Conference of the North {A}merican
  Chapter of the Association for Computational Linguistics: Human Language
  Technologies, Volume 1 (Long and Short Papers)}, pages 4171--4186.

\bibitem[{Fleiss(1971)}]{fleiss1971measuring}
Joseph~L Fleiss. 1971.
\newblock Measuring nominal scale agreement among many raters.
\newblock \emph{Psychological bulletin}, 76(5):378.

\bibitem[{Gu et~al.(2019)Gu, Cho, Ha, and Kim}]{gu2018dialogwae}
Xiaodong Gu, Kyunghyun Cho, Jung-Woo Ha, and Sunghun Kim. 2019.
\newblock Dialogwae: Multimodal response generation with conditional
  wasserstein auto-encoder.
\newblock In \emph{ICLR}.

\bibitem[{Gururangan et~al.(2019)Gururangan, Dang, Card, and
  Smith}]{gururangan2019variational}
Suchin Gururangan, Tam Dang, Dallas Card, and Noah~A Smith. 2019.
\newblock Variational pretraining for semi-supervised text classification.
\newblock In \emph{Proceedings of the 57th Annual Meeting of the Association
  for Computational Linguistics}, pages 5880--5894.

\bibitem[{Ke et~al.(2018)Ke, Guan, Huang, and Zhu}]{ke2018generating}
Pei Ke, Jian Guan, Minlie Huang, and Xiaoyan Zhu. 2018.
\newblock Generating informative responses with controlled sentence function.
\newblock In \emph{Proceedings of the 56th Annual Meeting of the Association
  for Computational Linguistics (Volume 1: Long Papers)}, pages 1499--1508.

\bibitem[{Li et~al.(2016)Li, Galley, Brockett, Gao, and
  Dolan}]{li2015diversity}
Jiwei Li, Michel Galley, Chris Brockett, Jianfeng Gao, and Bill Dolan. 2016.
\newblock A diversity-promoting objective function for neural conversation
  models.
\newblock In \emph{Proceedings of the 2016 Conference of the North {A}merican
  Chapter of the Association for Computational Linguistics: Human Language
  Technologies}, pages 110--119.

\bibitem[{Li et~al.(2020)Li, Feng, Wang, Song, Zhang, and
  Wang}]{li2021emoelicitor}
Shifeng Li, Shi Feng, Daling Wang, Kaisong Song, Yifei Zhang, and Weichao Wang.
  2020.
\newblock Emoelicitor: an open domain response generation model with user
  emotional reaction awareness.
\newblock In \emph{Proceedings of the Twenty-Ninth International Joint
  Conference on Artificial Intelligence}, pages 3637--3643.

\bibitem[{Li et~al.(2017)Li, Su, Shen, Li, Cao, and Niu}]{li2017dailydialog}
Yanran Li, Hui Su, Xiaoyu Shen, Wenjie Li, Ziqiang Cao, and Shuzi Niu. 2017.
\newblock Dailydialog: A manually labelled multi-turn dialogue dataset.
\newblock In \emph{Proceedings of the Eighth International Joint Conference on
  Natural Language Processing (Volume 1: Long Papers)}, pages 986--995.

\bibitem[{Miller et~al.(2017)Miller, Feng, Fisch, Lu, Batra, Bordes, Parikh,
  and Weston}]{miller2017parlai}
Alexander~H Miller, Will Feng, Adam Fisch, Jiasen Lu, Dhruv Batra, Antoine
  Bordes, Devi Parikh, and Jason Weston. 2017.
\newblock Parlai: A dialog research software platform.
\newblock In \emph{Proceedings of the 2017 Conference on Empirical Methods in
  Natural Language Processing: System Demonstrations}, pages 79--84.

\bibitem[{Roller et~al.(2020)Roller, Dinan, Goyal, Ju, Williamson, Liu, Xu,
  Ott, Smith, Boureau, and Weston}]{roller2020recipes}
Stephen Roller, Emily Dinan, Naman Goyal, Da~Ju, Mary Williamson, Yinhan Liu,
  Jing Xu, Myle Ott, Eric~Michael Smith, Y-Lan Boureau, and Jason Weston. 2020.
\newblock Recipes for building an open-domain chatbot.
\newblock In \emph{Proceedings of the 16th Conference of the European Chapter
  of the Association for Computational Linguistics: Main Volume}, pages
  300--325.

\bibitem[{Serban et~al.(2017)Serban, Sordoni, Lowe, Charlin, Pineau, Courville,
  and Bengio}]{serban2017hierarchical}
Iulian Serban, Alessandro Sordoni, Ryan Lowe, Laurent Charlin, Joelle Pineau,
  Aaron Courville, and Yoshua Bengio. 2017.
\newblock A hierarchical latent variable encoder-decoder model for generating
  dialogues.
\newblock In \emph{Proceedings of the Thirty-First {AAAI} Conference on
  Artificial Intelligence}, volume~31, pages 3295--3301.

\bibitem[{Vaswani et~al.(2017)Vaswani, Shazeer, Parmar, Uszkoreit, Jones,
  Gomez, Kaiser, and Polosukhin}]{vaswani2017attention}
Ashish Vaswani, Noam Shazeer, Niki Parmar, Jakob Uszkoreit, Llion Jones,
  Aidan~N. Gomez, Lukasz Kaiser, and Illia Polosukhin. 2017.
\newblock Attention is all you need.
\newblock In \emph{Annual Conference on Neural Information Processing Systems},
  pages 5998--6008.

\bibitem[{Vijayakumar et~al.(2016)Vijayakumar, Cogswell, Selvaraju, Sun, Lee,
  Crandall, and Batra}]{vijayakumar2016diverse}
Ashwin~K Vijayakumar, Michael Cogswell, Ramprasath~R Selvaraju, Qing Sun,
  Stefan Lee, David Crandall, and Dhruv Batra. 2016.
\newblock Diverse beam search: Decoding diverse solutions from neural sequence
  models.
\newblock \emph{arXiv preprint arXiv:1610.02424}.

\bibitem[{Wang and Wan(2019)}]{wang2019t}
Tianming Wang and Xiaojun Wan. 2019.
\newblock T-cvae: Transformer-based conditioned variational autoencoder for
  story completion.
\newblock In \emph{Proceedings of the Twenty-Eighth International Joint
  Conference on Artificial Intelligence}, pages 5233--5239.

\bibitem[{Wu et~al.(2020)Wu, Li, Wang, Chen, Wong, Feng, Huang, and
  Wang}]{wu2019guiding}
Bowen Wu, Mengyuan Li, Zongsheng Wang, Yifu Chen, Derek Wong, Qihang Feng,
  Junhong Huang, and Baoxun Wang. 2020.
\newblock Guiding variational response generator to exploit persona.
\newblock In \emph{Proceedings of the 58th Annual Meeting of the Association
  for Computational Linguistics}, pages 53--65.

\bibitem[{Zhang et~al.(2018)Zhang, Dinan, Urbanek, Szlam, Kiela, and
  Weston}]{zhang2018personalizing}
Saizheng Zhang, Emily Dinan, Jack Urbanek, Arthur Szlam, Douwe Kiela, and Jason
  Weston. 2018.
\newblock Personalizing dialogue agents: {I} have a dog, do you have pets too?
\newblock In \emph{Proceedings of the 56th Annual Meeting of the Association
  for Computational Linguistics (Volume 1: Long Papers)}, pages 2204--2213.

\bibitem[{Zhang et~al.(2020)Zhang, Sun, Galley, Chen, Brockett, Gao, Gao, Liu,
  and Dolan}]{zhang2019dialogpt}
Yizhe Zhang, Siqi Sun, Michel Galley, Yen-Chun Chen, Chris Brockett, Xiang Gao,
  Jianfeng Gao, Jingjing Liu, and Bill Dolan. 2020.
\newblock Dialogpt: Large-scale generative pre-training for conversational
  response generation.
\newblock In \emph{Proceedings of the 58th Annual Meeting of the Association
  for Computational Linguistics: System Demonstrations}, pages 270--278.

\bibitem[{Zhao et~al.(2017)Zhao, Zhao, and Eskenazi}]{zhao2017learning}
Tiancheng Zhao, Ran Zhao, and Maxine Eskenazi. 2017.
\newblock Learning discourse-level diversity for neural dialog models using
  conditional variational autoencoders.
\newblock In \emph{Proceedings of the 55th Annual Meeting of the Association
  for Computational Linguistics (Volume 1: Long Papers)}, pages 654--664.

\bibitem[{Zhou and Wang(2018)}]{zhou2017mojitalk}
Xianda Zhou and William~Yang Wang. 2018.
\newblock Mojitalk: Generating emotional responses at scale.
\newblock In \emph{Proceedings of the 56th Annual Meeting of the Association
  for Computational Linguistics (Volume 1: Long Papers)}, volume~1, pages
  1128--1137.

\end{thebibliography}
\bibliographystyle{acl_natbib}
\end{document}